\documentclass[sigconf,nonacm]{acmart}

\AtBeginDocument{%
  }

\setcopyright{acmlicensed}
\copyrightyear{2026}
\acmYear{2026}
\acmDOI{XXXXXXX.XXXXXXX}
\acmConference[Conference acronym 'XX]{Make sure to enter the correct
  conference title from your rights confirmation email}{2026}{City, Country}
\acmISBN{978-1-4503-XXXX-X/2026/XX}

\newcommand{\model}[1]{\texttt{#1}}

\begin{document}

\title{Can Open-Weight Large Language Models (LLMs) Simulate Human Survey Populations? A Cross-Instrument Calibration Study}

\author{Grandee Lee}
\affiliation{%
  \institution{Singapore University of Social Sciences}
  \city{Singapore}
  \country{Singapore}}
\email{grandeelee@suss.edu.sg}

\author{Wang Yue}
\affiliation{%
  \institution{Singapore University of Social Sciences}
  \city{Singapore}
  \country{Singapore}}
\email{wangyue@suss.edu.sg}

\renewcommand{\shortauthors}{Lee and Wang}

\begin{abstract}
Large language models (LLMs) are increasingly used to generate synthetic survey respondents and digital twins of real people, but whether their output preserves real human statistical structure, rather than surface plausibility, remains unresolved, and most existing evidence comes from proprietary models rather than open-weight ones. We evaluate three open-weight LLM families on a cross-instrument calibration task: conditioning personas on real respondents' verbatim answers to one psychometric instrument and measuring them on a second, construct-distance-controlled instrument, checked against a 2,058-person human panel. Across a 139-pair grid, the simulated cross-instrument correlation tracks the real human correlation at $r = 0.70$--$0.73$ in every model, driven mainly by correct sign rather than precise magnitude and concentrated in pairs of moderate construct distance. A correlation of this magnitude, obtained from untuned open-weight models conditioned only on individual-level survey data, is a substantively encouraging result for LLM-based behavioral simulation and digital-twin applications: specific model families and releases already reproduce a meaningful share of real human cross-instrument structure without any fine-tuning. This capability does not, however, improve monotonically across model releases: on a matched panel, the newest of three tested Llama releases performs worst on two of three headline metrics, so realizing its promise in practice requires release-specific, distance-aware verification rather than a one-time benchmark.
\end{abstract}



\maketitle

\section{Introduction}

Conditioning a language model on a persona and asking it to complete a survey instrument now produces simulated respondent data at negligible marginal cost, and this has been proposed as a substitute or complement for costly human data collection in the social and behavioral sciences~\cite{aher2023,argyle2023,bail2024,lee-etal-2026-generative}. The capability to produce such a population, however, is not evidence that its statistical properties can be trusted. Prior work using proprietary models finds that persona-conditioned populations reproduce plausible means while their response distributions collapse onto a handful of values~\cite{peng2026}, exaggerate the extremity of group attitudes relative to matched real respondents~\cite{bisbee2024}, and can match a target trait score without genuine inference, simply by echoing language present in the prompt or an instrument's own scoring key~\cite{shu2024,salecha2024}.

These findings raise two questions this paper addresses together for open-weight models specifically. First, \textit{does a model produce output consistent with genuine cross-construct inference, rather than repetition, when it is conditioned on one instrument and measured with a different one at varying theoretical and lexical distance?} Second, \textit{is whatever capability exists a stable property of a model family, or does it vary, potentially non-monotonically, across releases?} The second question has direct practical consequence: a pipeline validated against one release has no guarantee of remaining valid after an update, and our results show this concern is not hypothetical.

We test both questions on three open-weight models served through standard local inference tooling, so every result here is reproducible on commodity infrastructure without API access. Personas are conditioned on one real respondent's own verbatim item-level answers to a validated instrument, not a summarized trait score (we call the instrument used for conditioning the \emph{seed} instrument, and the one used for measurement the \emph{measure} instrument), and measured on a different instrument, with construct distance between the two controlled explicitly. Agreement between the resulting simulated cross-instrument correlation and the real human cross-instrument correlation for the same instrument pair is evidence of inference from the seed, and that evidence strengthens as the seed and measure instruments move further apart, both lexically and theoretically.

\paragraph{Contributions.}
\begin{enumerate}
  \item A cross-instrument calibration protocol that conditions LLM personas on real respondents' verbatim survey answers and checks the resulting simulated correlation against the real human panel, with an explicit construct-distance control. Simulated cross-instrument correlation tracks real human correlation ($r \approx 0.70$--$0.73$) across three independently trained open-weight model families.
  \item The first evidence, to our knowledge, that this calibration capability does not improve monotonically across successive releases of a model family: the newest of three tested Llama releases performs worst on two of three headline metrics.
  \item Two robustness checks on the conditioning mechanism itself: the simulated seed-to-measure correlation for a pair is not guaranteed to match its own reverse, and, on one model, a many-instrument seed generally out-calibrates a single instrument at recovering a held-out trait.
\end{enumerate}

\section{Related Work}

\begin{table*}[h!]
  \caption{Related work and its relation to this study.}
  \label{tab:related}
  \small\setlength{\tabcolsep}{3pt}
  \begin{tabular}{p{3.0cm}p{7.6cm}p{6.4cm}}
    \toprule
    Line of work & Representative findings & Relation to this study \\
    \midrule
    Individual-level survey conditioning & Conditioning a persona with a real respondent's own past answers, rather than demographic labels alone, substantially improves prediction of that person's held-out survey responses, up to a large share of a respondent's own test-retest reliability in the strongest reported cases~\cite{park2024,rupprecht2026} & We adopt individual-level, item-level conditioning as the design baseline, but target cross-instrument correlation \emph{structure} over a construct-distance spectrum, rather than per-item prediction accuracy on one instrument \\
    Digital twins, critically evaluated & Large-scale, richly profiled LLM ``digital twins'' show only weak correlation with held-out real human outcomes on broad outcome sets, alongside systematic distortions including insufficient individuation and ideological bias~\cite{peng2026} & Motivates testing whether population-level correlational structure can hold even where individual-level prediction is known to be limited \\
    Psychometric evaluation of LLMs & LLMs recover a prompted trait on the \emph{same} instrument with large effect sizes, but display social-desirability inflation, compressed response variance, and sensitivity to superficial prompt changes~\cite{serapio2025,salecha2024,shu2024,petrov2024,jiang2024} & We measure on a \emph{different} instrument from the one used as the seed, which is specifically designed to rule out same-instrument echo as an explanation for any observed agreement \\
    Fidelity and distributional collapse & Simulated populations reproduce plausible aggregate means while their underlying response distributions and cross-group variance collapse relative to real respondents~\cite{bisbee2024,santurkar2023,hwang2023,ozkan2026} & Motivates checking simulated correlation \emph{magnitude}, not only its sign, against the real human value \\
    Belief-action consistency & Stated preferences and coded downstream choices by the same LLM-simulated agent frequently diverge~\cite{shen2025} & We include one small-scale behavioral extension in this spirit but treat it as preliminary (Section~\ref{sec:behavior}), using the choice task of Cherep et al.~\cite{cherep2026} \\
    \bottomrule
  \end{tabular}
\end{table*}

Table~\ref{tab:related} summarizes the lines of work this study builds on and how it relates to each.

To our knowledge, no prior open-weight evaluation combines individual-level survey conditioning, an explicit construct-distance control, and a multi-model, multi-release comparison in one design (cf.\ Lutz et al.~\cite{lutz2025}, on sociodemographic rather than individual-level persona prompting). The design separates three questions: whether agreement exists, whether surface repetition explains it, and whether it is stable across models and releases.

\section{Method}

\subsection{Human reference panel}

We use a national panel of 2,058 respondents~\cite{toubia2025} with complete responses to a battery of psychometric instruments covering personality (a Big Five inventory, scored as five separate, largely independent facet scores rather than a single composite), cognitive style, values and economic preference orientations, social and interpersonal disposition, and self-presentation bias. Fifteen instruments have complete, independently verified item-level coverage sufficient to serve as either a seed or a measurement target; one additional candidate instrument was excluded because its ground-truth coverage was incomplete.

\subsection{Persona conditioning}

A simulated persona is constructed from one real panel respondent's verbatim, item-level answers to a single instrument (for example, ``When asked `[item text]' you responded `[response label]''' for every item of that instrument), never from a summarized trait score. The model is then asked, in a fresh context, to complete a \emph{different} instrument under a response-format schema constrained to that instrument's own scale. This is repeated for 40 real respondents per seed-measure pair, drawn reproducibly from the pool of respondents common to both instruments in the pair.

\subsection{Construct-distance control}
\label{sec:distance}

Every unordered pair of the fifteen usable instruments is scored on two independent axes, computed by a fixed procedure rather than assigned ad hoc: a lexical-overlap score (maximum item-level Jaccard token similarity) and a four-tier construct-distance judgment (a family-based default, overridden for specific pairs with a citable relationship in the literature). Appendix~\ref{app:distance} gives the full procedure, family assignments, and overrides. We treat this as a defensible, literature-grounded starting point, not an independently validated distance metric: a result at low lexical overlap and a high tier is harder to attribute to surface pattern-matching, and is correspondingly stronger evidence of genuine cross-construct inference.

\subsection{Calibration criterion}

A calibrated simulated population should reproduce the real human cross-instrument correlation for a given pair. We assess this at two levels, defined in Section~\ref{sec:grid}: population-level agreement between the simulated and human seed-measure correlations across pairs (Eqs.~\ref{eq:pearson}--\ref{eq:mae}), and respondent-level agreement between each persona's simulated measure score and that same respondent's real score.

\subsection{Models and inference settings}

Three open-weight models were served through standard local inference infrastructure: \model{gemma4:31b}, \model{llama3.1:70b}, and \model{qwen3.8:27b}, each evaluated at its own documented default sampling temperature (1.0, 0.8, and 0.8 respectively) unless a temperature sweep is explicitly described. A separate comparison (Section~\ref{sec:releases}) additionally evaluates two further releases within the Llama family, \model{llama3:70b} and \model{llama3.3:70b}.


\section{Results}

\subsection{Necessity of conditioning}
\label{sec:seeding}

\begin{figure*}[ht!]
  \centering
  \includegraphics[width=0.75\textwidth]{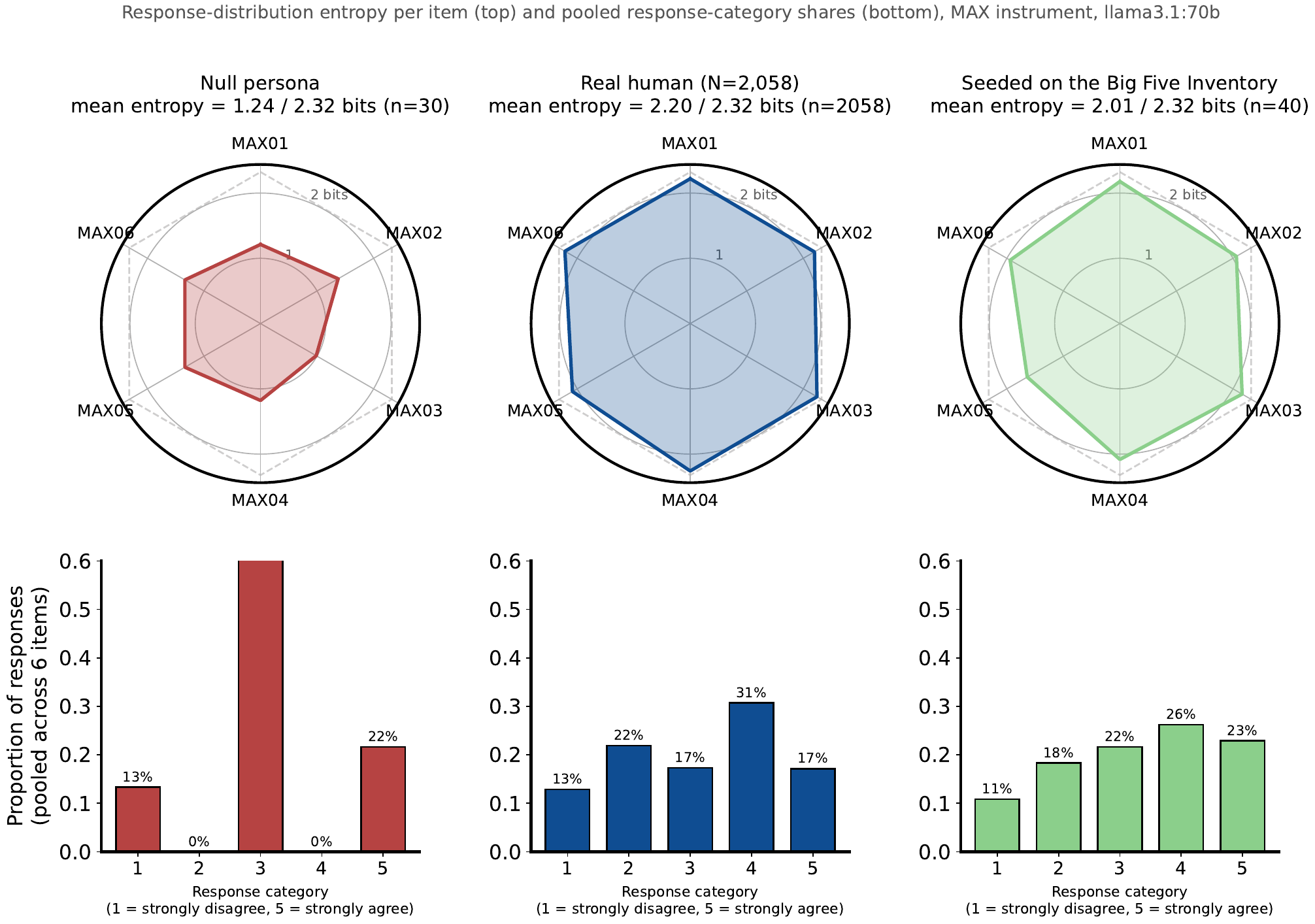}
  \caption{Top: response-distribution entropy per item, Maximization Scale (MAX)~\cite{nenkov2008}, \model{llama3.1:70b}, null persona (left), real human population (center), conditioned on the Big Five Inventory (right). Bottom: the same three conditions' pooled response-category shares across all six items.}
  \Description{Three radar (hexagon) plots of per-item response entropy for the null persona, the real human population and the Big Five-conditioned persona, above three bar charts of pooled response-category shares for the same conditions.}
  \label{fig:radar}
\end{figure*}

An unconditioned (``null'') persona's item-level response distribution collapses onto a small number of response categories, regardless of sampling temperature, while conditioning on a single instrument restores use of the full response scale on nearly every item of a held-out measure (Figure~\ref{fig:radar}). Each hexagon axis is one of the six items of the Maximization Scale (MAX)~\cite{nenkov2008}, and the radius is the entropy of that item's responses in bits, from 0 (every response identical) to 2.32 (all five categories equally likely, the dashed circle). Entropy measures ``collapse'' better than variance here: the null persona splits its answers between the scale's endpoints and midpoint, skipping categories 2 and 4 on every item and placing 60\% of responses on the midpoint, yet keeps a variance comparable to the other conditions. Its mean entropy is 1.24 bits, against 2.20 for the real population; conditioning on a single instrument recovers most of the shortfall (2.01 bits) with no change to temperature. As Section~\ref{sec:temp} shows, raising temperature never produces a comparable improvement, so it is no substitute for conditioning.

\subsection{Effect of sampling temperature}
\label{sec:temp}

We swept each model's own default temperature plus two upward increments (+0.2, +0.4) against six instrument pairs ($n = 40$ per pair), measuring the mean absolute error (MAE, Eq.~\ref{eq:mae}) between the simulated and real cross-instrument correlation (Table~\ref{tab:temp}). All three models deviate further from the human correlation as temperature rises above their own default, without exception.

\begin{table}[h!]
  \caption{MAE between simulated and real cross-instrument correlation, by sampling temperature.}
  \label{tab:temp}
  \begin{tabular}{lccc}
    \toprule
    Model & Own default & +0.2 & +0.4 \\
    \midrule
    \model{gemma4:31b}   & 0.406 & 0.461 & 0.473 \\
    \model{llama3.1:70b} & 0.254 & 0.334 & 0.363 \\
    \model{qwen3.8:27b}  & 0.228 & 0.275 & 0.314 \\
    \bottomrule
  \end{tabular}
\end{table}

\subsection{Cross-instrument calibration across three open-weight models}
\label{sec:grid}

We extended the single-instrument-seed grid to all ordered pairs among the fifteen instruments not already covered by the temperature sweep, run identically across all three models (same 40 respondents per pair, each model's own default temperature), giving $P = 139$ pairs per model. Each run is an ordered pair $p = (a{\to}b)$ of a seed instrument $a$ and a measure instrument $b$. For respondent $i$, let $s_i^{a}$ be their real score on instrument $a$ and $\hat{s}_i^{a\to b}$ the score on $b$ produced by their persona conditioned on $a$. Two quantities are evaluated.

\emph{Seed-measure correlation, human vs.\ simulated (Table~\ref{tab:grid}).} For each pair we compute the human seed-measure correlation $\rho^{H}_p = \mathrm{corr}_i(s_i^{a}, s_i^{b})$ (from the full 2,058-person panel) and the simulated one $\rho^{S}_p = \mathrm{corr}_i(s_i^{a}, \hat{s}_i^{a\to b})$. The columns of Table~\ref{tab:grid} compare these two quantities across the $P$ pairs:
\begin{equation}
  \text{Pearson } r = \mathrm{corr}_p\!\left(\rho^{S}_p,\, \rho^{H}_p\right), \label{eq:pearson}
\end{equation}
\begin{align}
  \text{Spearman } r_s &= \mathrm{corr}_p\!\left(\mathrm{rank}\,\rho^{S}_p,\, \mathrm{rank}\,\rho^{H}_p\right), \label{eq:spearman} \\
  \text{Sign agr.} &= \frac{1}{P}\sum_{p} \mathbb{1}\!\left[\mathrm{sgn}\,\rho^{S}_p = \mathrm{sgn}\,\rho^{H}_p\right], \label{eq:sign}\\
  \text{MAE} &= \frac{1}{P}\sum_{p} \left|\rho^{S}_p - \rho^{H}_p\right|. \label{eq:mae}
\end{align}

\emph{Respondent-level agreement, simulated vs.\ real measure score.} For each pair, $v_p = \mathrm{corr}_i(s_i^{b}, \hat{s}_i^{a\to b})$ asks whether the persona conditioned on a real person's seed answers reproduces that same person's own answer to the measure instrument, regardless of whether the pair-level relationship replicates. Averaged over pairs, $\bar{v} = \frac{1}{P}\sum_p v_p$ is $0.210$ for \model{gemma4:31b}, $0.203$ for \model{llama3.1:70b}, and $0.152$ for \model{qwen3.8:27b} (medians $0.194$, $0.192$, $0.098$; $v_p$ is negative for 23, 22, and 28 of the 139 pairs). Respondent-level agreement from a single seed instrument is therefore modest, well below the population-level agreement of Table~\ref{tab:grid}; Section~\ref{sec:loo} shows how it changes with a many-instrument seed.

\begin{table}[h!]
  \caption{Agreement between simulated ($\rho^{S}_p$) and real human ($\rho^{H}_p$) seed-measure correlations across the 139-pair grid; all statistics are computed across pairs.}
  \label{tab:grid}
  \small\setlength{\tabcolsep}{3pt}
  \resizebox{\columnwidth}{!}{%
\begin{tabular}{lccccc}
    \toprule
    Model & $n$ & Pearson $r$ & Spearman $r_s$ & Sign agr. & MAE \\
    \midrule
    \model{gemma4:31b}   & 139 & 0.703 & 0.628 & 87.1\% & 0.276 \\
    \model{llama3.1:70b} & 139 & 0.731 & 0.671 & 86.3\% & 0.316 \\
    \model{qwen3.8:27b}  & 139 & 0.701 & 0.640 & 82.7\% & 0.194 \\
    \bottomrule
  \end{tabular}}
\end{table}

Against the same-sized ($n = 40$) human sample instead of the full panel, the Pearson agreement is weaker (0.530, 0.557, 0.568), partly because the small human reference is itself noisy: its own $n = 40$ and full-population correlations agree only at $r = 0.809$. The four construct-distance tiers of Section~\ref{sec:distance} differ greatly in size and composition (Table~\ref{tab:tiers}).

\begin{table}[h!]
  \caption{Composition of the four construct-distance tiers in the 139-pair grid.}
  \label{tab:tiers}
  \small\setlength{\tabcolsep}{3pt}
  \resizebox{\columnwidth}{!}{%
\begin{tabular}{lcccc}
    \toprule
    Tier & Pairs & Distinct seeds & Human corr.\ range & Human corr.\ SD \\
    \midrule
    1 (near-paraphrase)         & 6  & 2  & [$-0.41$, 0.72] & 0.339 \\
    2                           & 52 & 13 & [$-0.56$, 0.62] & 0.225 \\
    3                           & 75 & 13 & [$-0.23$, 0.37] & 0.146 \\
    4 (theoretically unrelated) & 6  & 4  & [0.00, 0.15]    & 0.064 \\
    \bottomrule
  \end{tabular}}
\end{table}

\begin{figure*}[h!]
  \centering
  \includegraphics[width=\textwidth]{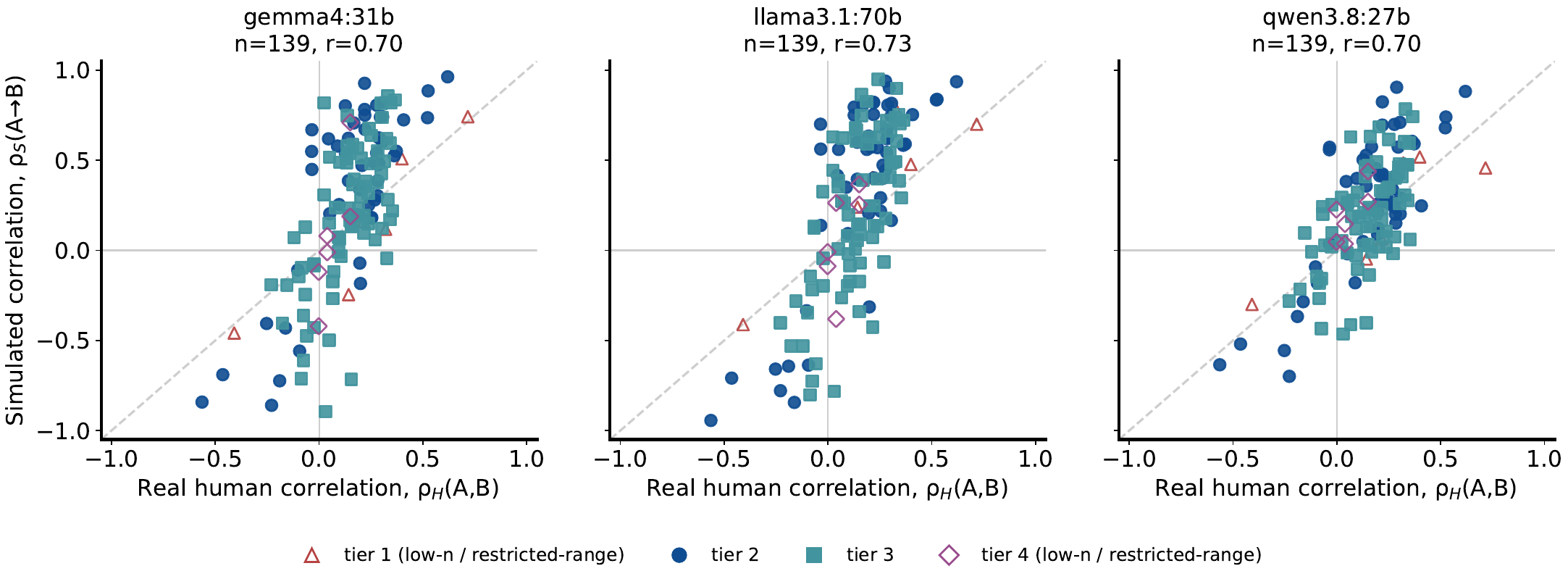}
  \caption{Simulated seed-to-measure correlation plotted against real human cross-instrument correlation for all 139 pairs, one panel per model, points shaped and colored by construct-distance tier.}
  \Description{Three scatter plots, one per model, of simulated versus human correlation with an identity line; markers differ by construct-distance tier, with tier 1 and tier 4 drawn hollow.}
  \label{fig:scatter}
\end{figure*}

Tier 1's six pairs draw on only two distinct seed instruments, so they are far fewer than six independent tests, and tier 4's human correlations sit in a narrow band near zero, so a low error there is easy to obtain. Only tiers 2 and 3 are large and diverse enough to interpret on their own; tiers 1 and 4 are retained but flagged (hollow markers in Figure~\ref{fig:scatter}). Overall, agreement holds across all three model families in a narrow 0.70--0.73 band; \model{qwen3.8:27b} has the lowest magnitude error (0.194) but also the lowest sign agreement (82.7\%).

\subsection{Effect of construct distance}
\label{sec:tiers}

\begin{table*}
  \caption{Agreement between simulated and human seed-measure correlations by construct-distance tier: $r$ is the Pearson correlation and MAE the mean absolute difference between them, across the pairs in the tier. The last three columns give $r$ (and MAE for all pairs) after excluding tier 1, restricting to below-median lexical overlap, and using all pairs.}
  \label{tab:bytier}
  \resizebox{\textwidth}{!}{%
\begin{tabular}{lccccccc}
    \toprule
    Model & Tier 1 ($r$ / MAE) & Tier 2 ($r$ / MAE) & Tier 3 ($r$ / MAE) & Tier 4 ($r$ / MAE) & Excl.\ tier 1 & Low lexical overlap only & All pairs \\
    \midrule
    \model{gemma4:31b}   & 0.92 / 0.14 & 0.78 / 0.32 & 0.63 / 0.26 & 0.85 / 0.20 & 0.71 & 0.69 & 0.70 / 0.28 \\
    \model{llama3.1:70b} & 0.91 / 0.14 & 0.81 / 0.38 & 0.66 / 0.30 & 0.64 / 0.18 & 0.74 & 0.72 & 0.73 / 0.32 \\
    \model{qwen3.8:27b}  & 0.87 / 0.17 & 0.79 / 0.21 & 0.59 / 0.19 & 0.75 / 0.13 & 0.71 & 0.72 & 0.70 / 0.19 \\
    \bottomrule
  \end{tabular}}
\end{table*}

Within each tier we compute the Pearson $r$ (Eq.~\ref{eq:pearson}) and MAE (Eq.~\ref{eq:mae}) over only the pairs $p$ in that tier, rather than over all $P$ pairs (Table~\ref{tab:bytier}). Within the two reliably sized strata, $r$ declines with construct distance in every model (0.78--0.81 at tier 2 to 0.59--0.66 at tier 3). MAE does not track distance the same way: in all three models it is highest at tier 2, not at the more distant tier 3, and falls further at tier 4. Excluding the six tier-1 pairs, or restricting to below-median lexical overlap, changes the aggregate $r$ by no more than 0.03 in any model. Because tier 1 is only 4\% of the grid, this check has limited power to detect leakage, and should be read as not contradicting a genuine-inference account rather than as strong confirmation of one.

\subsection{Directionality of seed-measure pairs}
\label{sec:direction}

The human correlation between two instruments is symmetric, but the simulated one need not be. Conditioning on instrument $a$ and generating $b$ is a different computation from conditioning on $b$ and generating $a$. In the notation of Section~\ref{sec:grid}, the simulated correlation for the forward direction is estimated across respondents as
\begin{equation}
  \rho^{S}_{a\to b} \approx \mathrm{corr}_{i}\!\left(s_i^{a},\, \hat{s}_i^{a\to b}\right),
\end{equation}
for a given model and temperature. Because the two directions come from different generative processes,
\begin{equation}
  \rho^{S}_{a\to b} \neq \rho^{S}_{b\to a} \text{ in general.}
\end{equation}
We measure this disagreement with $\Delta_{ab} = \rho^{S}_{a\to b} - \rho^{S}_{b\to a}$ on every pair run in both directions among the fourteen instruments outside the Big Five (27 pairs shared by all three models; Table~\ref{tab:asym}). The Big Five is excluded because it is always conditioned as a full profile but measured per facet, so it has no true reverse.

\begin{table}
  \caption{Forward vs.\ reverse simulated correlation for bidirectional pairs.}
  \label{tab:asym}
  \small\setlength{\tabcolsep}{3pt}
  \resizebox{\columnwidth}{!}{%
\begin{tabular}{lccccc}
    \toprule
    Model & Pairs & $\mathrm{corr}(\rho^{S}_{a\to b}, \rho^{S}_{b\to a})$ & Mean $|\Delta|$ & Mean signed $\Delta$ & Max $|\Delta|$ \\
    \midrule
    \model{gemma4:31b}   & 27 & 0.666 & 0.247 & $-0.062$ & 0.666 \\
    \model{llama3.1:70b} & 27 & 0.718 & 0.237 & $-0.011$ & 0.644 \\
    \model{qwen3.8:27b}  & 27 & 0.139 & 0.260 & $+0.020$ & 1.042 \\
    \bottomrule
  \end{tabular}}
  \\[2pt]
  \footnotesize $\Delta_{ab}$ is the difference $\rho^{S}_{a\to b} - \rho^{S}_{b\to a}$.
\end{table}

\begin{figure*}
  \centering
  \includegraphics[width=\textwidth]{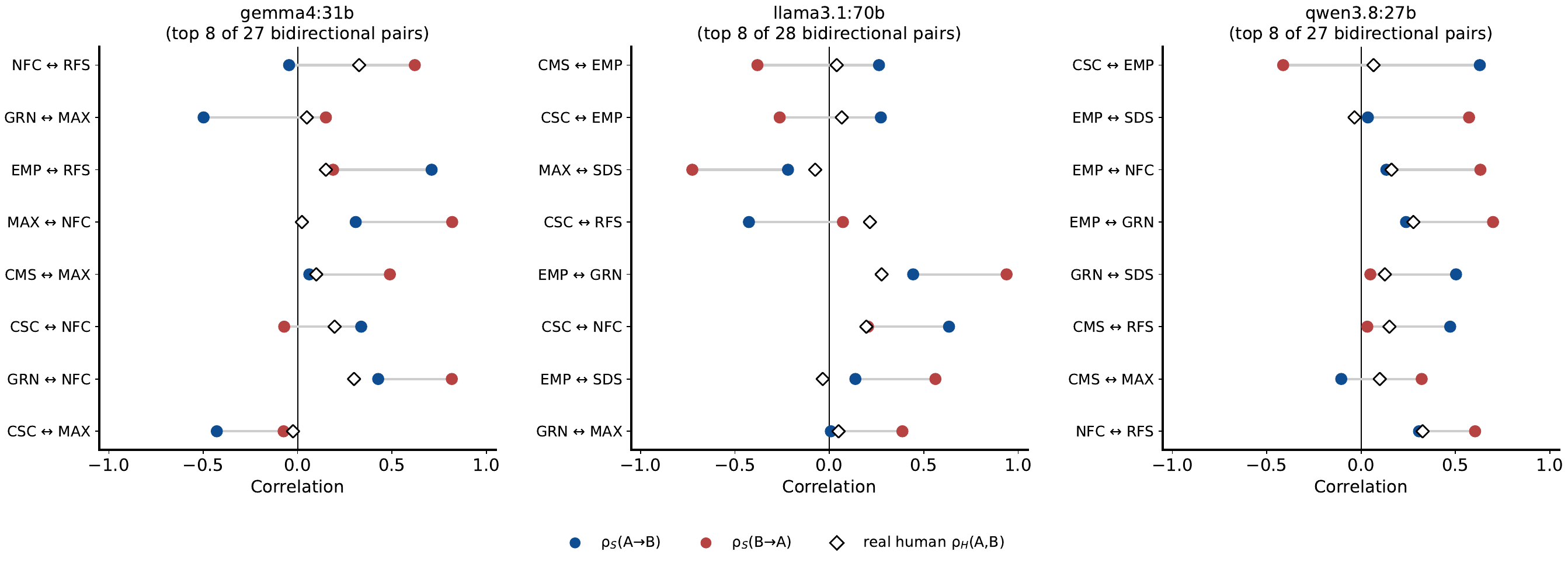}
  \caption{The pairs with the largest directional difference between $\rho^{S}_{a\to b}$ and $\rho^{S}_{b\to a}$, one panel per full-grid model, with the real human correlation marked for reference.}
  \Description{One panel per model showing, for the pairs with the largest forward-versus-reverse difference, both simulated correlations alongside the real human correlation.}
  \label{fig:asym}
\end{figure*}

Three findings stand out. First, forward and reverse correlate at 0.67--0.72 across pairs for \model{gemma4:31b} and \model{llama3.1:70b}, but only 0.139 for \model{qwen3.8:27b}, whose largest swing (Conscientiousness conditioned on Empathy, $+0.629$, against the reverse, $-0.413$) flips sign depending only on which instrument is the seed. Second, mean signed $\Delta$ is near zero in every model, so the asymmetry is pair-specific rather than a bias toward the first-named instrument; it concentrates on pairs whose real correlation is near zero, and the two directions often land on opposite sides of the human value (Figure~\ref{fig:asym}), so a result reported in one direction alone is one noisy sample of the pair. Third, the Big Five as a seed anchors the \emph{direction} of a relationship more reliably than as a measure: Big-Five-conditioned pairs get the sign right on all eleven pairs in every model, against 81--90\% sign agreement when a Big Five facet is the measure, though eleven pairs per model is a small, unmatched sample.

\subsection{Combining seed instruments: leave-one-out validity}
\label{sec:loo}

Does conditioning on many instruments at once let the model recover a specific person's own trait, not just the right population-level relationship? We test this on one model, \model{llama3.1:70b}, the only one on which a many-instrument seed has been run. For each of nineteen held-out targets $t$ (the fourteen general instruments plus the Big Five's five facets), we condition a persona on one real respondent's verbatim answers to the \emph{other} fourteen instruments, measure it fresh on $t$, and do this for forty respondents, each conditioned on their own profile. The criterion is the respondent-level agreement of Section~\ref{sec:grid}, $v_t = \mathrm{corr}_i(s_i^{t}, \hat{s}_i^{\text{-}t\to t})$, where $\hat{s}_i^{\text{-}t\to t}$ is the score on target $t$ produced by the persona conditioned on all other instruments: the correlation across respondents between the simulated and real score on the held-out target.

Validity is positive for all nineteen targets (mean $v = 0.424$, median $0.406$, SD $0.193$), and sixteen exceed $0.3$. The strongest are Conscientiousness ($0.817$) and Openness ($0.709$); the weakest is the Maximization Scale ($0.032$). No target is negative, so the model does not systematically invert a held-out trait, even where tracking is weak.

\begin{figure}
  \centering
  \includegraphics[width=\linewidth]{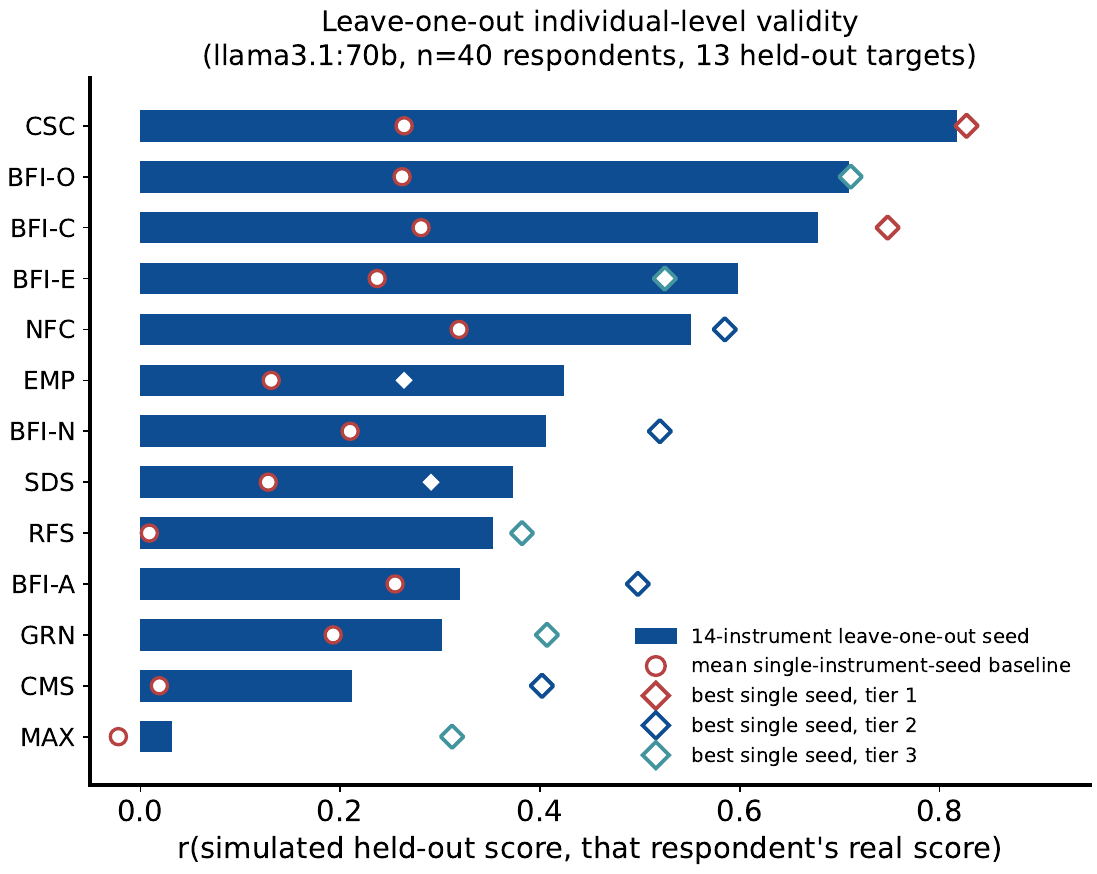}
  \caption{Respondent-level leave-one-out validity ($v_t$), one bar per held-out target, sorted descending, with mean and best available single-instrument-seed baselines overlaid; the best-single-seed marker is colored by that specific seed-measure pair's construct-distance tier.}
  \Description{Bar chart of individual-level correlation for each of nineteen held-out targets, with markers for the mean single-instrument baseline and the best single-instrument seed.}
  \label{fig:loo}
\end{figure}

\begin{table}
  \caption{Leave-one-out (LOO) joint-seed respondent-level validity ($v$) vs.\ mean single-instrument-seed validity.}
  \label{tab:loo}
  \small\setlength{\tabcolsep}{3pt}
  \resizebox{\columnwidth}{!}{%
\begin{tabular}{lcccc}
    \toprule
    Measure & 14-instr.\ LOO $v$ & Mean single-instr.\ $v$ & $n$ single-seed comp. & Diff. \\
    \midrule
    BFI-A & 0.320 &  0.255 & 14 & $+0.065$ \\
    BFI-C & 0.678 &  0.281 & 13 & $+0.397$ \\
    BFI-E & 0.598 &  0.237 & 14 & $+0.361$ \\
    BFI-N & 0.406 &  0.210 & 14 & $+0.196$ \\
    BFI-O & 0.709 &  0.262 & 14 & $+0.447$ \\
    NFC   & 0.551 &  0.319 &  8 & $+0.232$ \\
    CMS   & 0.212 &  0.019 &  8 & $+0.193$ \\
    EMP   & 0.424 &  0.131 &  8 & $+0.293$ \\
    GRN   & 0.302 &  0.193 &  8 & $+0.109$ \\
    CSC   & 0.817 &  0.264 &  8 & $+0.553$ \\
    SDS   & 0.373 &  0.128 &  8 & $+0.245$ \\
    RFS   & 0.353 &  0.009 &  8 & $+0.344$ \\
    MAX   & 0.032 & $-0.022$ & 8 & $+0.054$ \\
    \bottomrule
  \end{tabular}}
\end{table}

We compare each target's $v_t$ with the mean $v$ over every single-instrument-seed run for the same target (Table~\ref{tab:loo}), restricted to the thirteen targets from the nine most thoroughly verified instruments, each with eight to fourteen single-seed runs. All thirteen targets improve under the joint seed (mean difference $+0.268$), from CSC ($0.264 \to 0.817$) to the smallest gain, MAX ($-0.022 \to 0.032$). This is a between-run comparison, not a paired one, since the runs do not share the same seed content, so it suggests that seed depth matters without isolating it as the cause. Against the single \emph{best} seed instrument for each target rather than the mean, the picture is mixed: the joint seed wins on three of thirteen targets (EMP, SDS, BFI-E), while the best single instrument wins on the other ten (for example CSC, $0.827$ vs.\ $0.817$; BFI-A, $0.498$ vs.\ $0.320$). A many-instrument seed is therefore a more reliable \emph{default} than an arbitrary single instrument, but does not generally beat the best single instrument when it is known in advance. The two highest best-single-seed values (CSC and BFI-C) come from tier-1 pairs, consistent with Section~\ref{sec:tiers}; the other eleven are tier 2 or 3, with no clear ordering between them.

\subsection{Calibration across successive model releases}
\label{sec:releases}

Whether calibration improves as a single family is updated determines whether a result obtained on one release can be assumed to hold after an update. We tested three successive 70-billion-parameter Llama releases (\model{llama3:70b}, \model{llama3.1:70b}, \model{llama3.3:70b}) on a fixed, stratified 14-pair panel (two pairs from each of the seven tier-by-lexical-overlap strata), with \model{gemma4:31b} and \model{qwen3.8:27b} as reference points (Table~\ref{tab:releases}). None of the three metrics improves monotonically across releases: \model{llama3.1:70b} has the best correlation and lowest MAE of the three, while the newest, \model{llama3.3:70b}, has the worst of both and a markedly lower sign agreement (71.4\% vs.\ 92.9\% for each older release). A 14-pair panel carries substantial sampling variability (\model{qwen3.8:27b} scores 82.7\% sign agreement on the full grid but 71.4\% here), so the exact ranking is indicative only. The absence of a monotonic pattern is the relevant finding: it argues against assuming that a newer release of the same family improves this capability without direct verification.

\begin{table}
  \caption{Calibration metrics on the fixed 14-pair panel across Llama releases.}
  \label{tab:releases}
  \small\setlength{\tabcolsep}{3pt}
  \begin{tabular}{lcccc}
    \toprule
    Model & $n$ & Pearson $r$ & Sign agr. & MAE \\
    \midrule
    \model{llama3:70b}                & 14 & 0.679 & 92.9\% & 0.383 \\
    \model{llama3.1:70b}              & 14 & 0.696 & 92.9\% & 0.320 \\
    \model{llama3.3:70b}              & 14 & 0.649 & 71.4\% & 0.403 \\
    \model{gemma4:31b} (reference)    & 14 & 0.628 & 85.7\% & 0.370 \\
    \model{qwen3.8:27b} (reference)   & 14 & 0.701 & 71.4\% & 0.207 \\
    \bottomrule
  \end{tabular}
\end{table}

\subsection{Extending calibration to a behavioral task}
\label{sec:behavior}

We extended the conditioning procedure to the default-option choice task of Cherep et al.~\cite{cherep2026}, in which simulated participants choose among options under a default-option nudge, using one real respondent's profile as conditioning material, added cumulatively across up to four instruments (personality, cognitive style, regulatory focus, and a decision-style measure). For \model{llama3.1:70b}, default-nudge acceptance rose from 0.68 under a null persona to 0.84 under the full four-instrument seed, against a human baseline of 0.88; the no-nudge control condition's baseline-choice rate rose from 0.25 toward 0.51, matching the human baseline. Two synthetic personas constructed to be nudge-resistant and nudge-susceptible, respectively, by manipulating two of the four seed instruments toward literature-predicted extremes, moved in the predicted direction (0.56 versus 0.74 default-nudge acceptance) without exceeding the real-respondent four-instrument result.

\begin{figure}
  \centering
  \includegraphics[width=\linewidth]{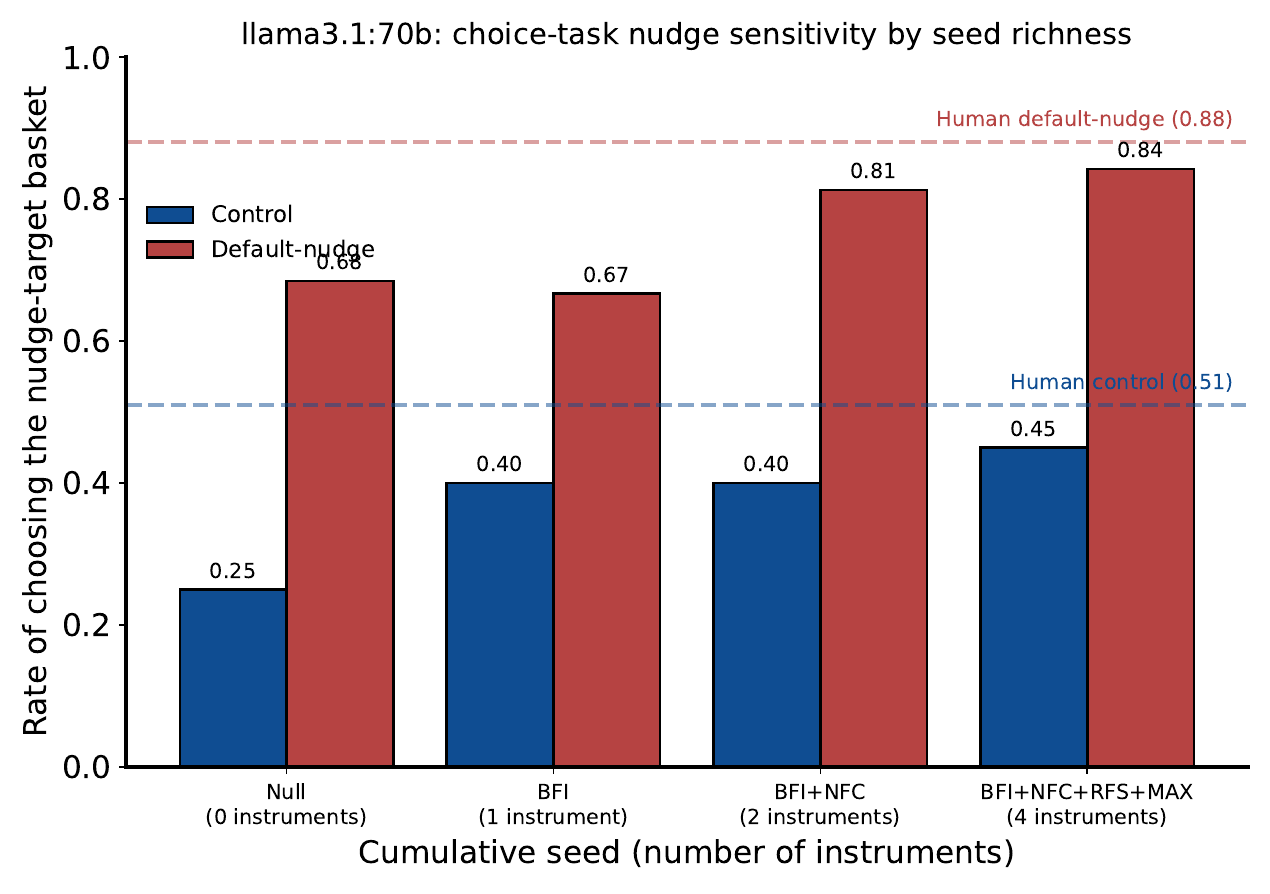}
  \caption{Choice-task control and default-nudge acceptance rates, paired bars at each of the four cumulative seed-richness levels, against human control and default-nudge baselines.}
  \Description{Paired bar chart of control and default-nudge acceptance rates at four cumulative conditioning levels, with horizontal reference lines for the human baselines.}
  \label{fig:dose}
\end{figure}

\begin{figure}
  \centering
  \includegraphics[width=\linewidth]{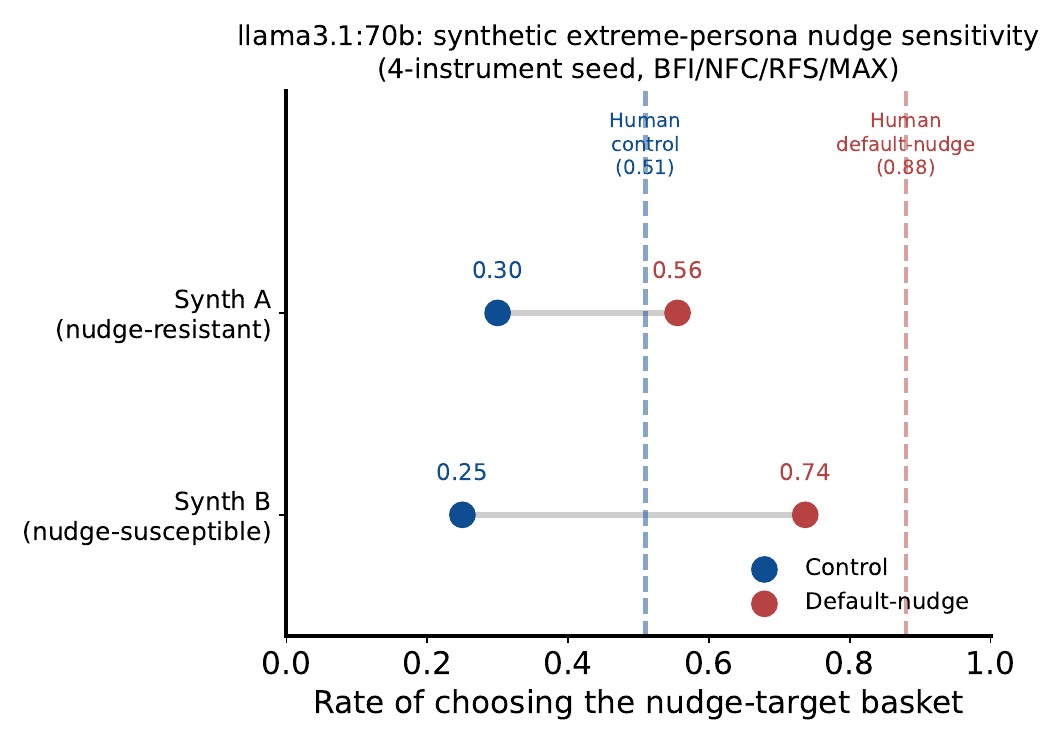}
  \caption{Synthetic extreme-persona nudge sensitivity, one dumbbell per persona connecting its control and default-nudge rate, against human control and default-nudge baselines.}
  \Description{Dumbbell plot with one row per synthetic persona connecting its control-condition rate to its default-nudge rate, with human baselines marked.}
  \label{fig:extreme}
\end{figure}

We flag this section as preliminary. These multi-instrument seeds were constructed and run prior to a correction ensuring the full seed text is reliably preserved within the model's input context on every call; because the longest seeds used here approach the context length at which truncation becomes possible, we report these results for completeness but do not treat them as confirmed until independently re-verified. They are limited in any case to a single model, a single real respondent, and one behavioral task.

\section{Discussion}

Four findings characterize what current open-weight models can do on this task. First, cross-instrument calibration is a replicable phenomenon: three independently trained families converge on $r \approx 0.70$--$0.73$ between simulated and human seed-measure correlations, from nothing more than one instrument's worth of individual-level conditioning, and the result is not carried by the few pairs most open to a surface-matching explanation. Second, the models predict the \emph{sign} of a relationship better than its \emph{magnitude}, and magnitude error does not fall with construct distance the way correlation does, so a practitioner can treat direction as trustworthy and magnitude as approximate. Third, the capability does not track release recency: the newest of three Llama releases is worst on two of three metrics, consistent with earlier evidence that size and recency do not reliably predict persona-simulation quality~\cite{peng2026}, here extended to a within-family, cross-release comparison. Fourth, a pair's simulated correlation need not agree with its own reverse (0.67--0.72 for two models, 0.14 for the third), which bounds the weight any single seed-measure direction deserves.

A simple frame unifies several of these results, though none was designed to test it. An untuned model encodes a broad superposition over the respondents compatible with its training data, and left unconditioned it collapses onto a narrow, unrepresentative slice of the response scale (Section~\ref{sec:seeding}). A verbatim, individual-level seed narrows that superposition toward a specific person, and richer seed material should narrow it more. This fits the Big Five anchoring direction better as a seed than as a measure (Section~\ref{sec:direction}) and, on the one model tested, a fourteen-instrument seed generally out-calibrating a single instrument (Section~\ref{sec:loo}). Projecting a rich profile forward onto an unseen instrument is then an easier act of narrowing than inferring the profile backward from one different instrument, so directional asymmetry follows from the difference in difficulty and is not a separate phenomenon.

Overall, the results support a constructive stance on open-weight LLMs as simulated survey respondents: the capability is real, replicable across model families, and substantial enough to found behavioral-simulation and digital-twin pipelines built without proprietary models. It is also uneven across construct distance and cannot be assumed to carry over from one release to the next, so calibration of this kind should be an ongoing, release-specific and distance-aware check, not a one-time validation. That overhead is modest compared with collecting new human data.

\section{Limitations and Future Directions}

The 139 pairs per model share the same 40 respondents and reuse instruments, so the reported statistics are descriptive, not supported by independence-assuming significance tests; the 27 bidirectional pairs (Section~\ref{sec:direction}) inherit this and are too few to test whether \model{qwen3.8:27b}'s weaker forward/reverse agreement is stable. No non-LLM baseline (for example a regression predictor or a generative sampler fit to the human panel) is reported, so $r \approx 0.70$--$0.73$ cannot be judged against what a simpler method achieves. The two smallest construct-distance strata (Section~\ref{sec:tiers}) are low in seed diversity or restricted in range and should not be extrapolated. The behavioral extension (Section~\ref{sec:behavior}) covers one model and one respondent, with provisional results; the release comparison (Section~\ref{sec:releases}) covers one family on a small panel; and the leave-one-out result (Section~\ref{sec:loo}) is a single-model study on \model{llama3.1:70b}, comparing only thirteen targets because the six newer instruments have too few single-seed runs. Finally, models are identified by public release tag, so exact reproducibility at the level of weights would require checksums we did not collect.

Future work should establish a non-LLM baseline for the same grid, extend the release comparison to more families and a larger panel with verified model identity, explain why \model{qwen3.8:27b}'s forward and reverse correlations agree less than the others', and re-verify the behavioral extension before treating it as more than a preliminary demonstration.

\section{Conclusion}

We evaluated whether three open-weight language models, conditioned on real individuals' survey responses, reproduce human cross-instrument correlational structure across a construct-distance-controlled grid, and whether this holds across successive releases of one family. Conditioning is necessary, and a model's own default temperature is preferable to any higher setting tested. Simulated seed-measure correlation tracks the human one at $r \approx 0.70$--$0.73$ in all three families, driven more by correctly signed relationships than by precisely estimated magnitudes, and not explained by the few near-paraphrase pairs. This capability does not, however, improve monotonically across three Llama releases: the newest performs worst on two of three metrics. Current open-weight models thus offer a substantial, encouraging foundation for behavioral-simulation and digital-twin applications without fine-tuning or proprietary infrastructure, with \model{llama3.1:70b} and \model{qwen3.8:27b} the most consistently calibrated in our tests. Because the capability is uneven across construct distance and not guaranteed to persist across updates, it should be verified release by release, not validated once and assumed.

\bibliographystyle{ACM-Reference-Format}
\bibliography{references}

\appendix

\section{Instruments Used}
\label{app:instruments}

\begin{table}
  \caption{The fifteen instruments used, with their sources.}
  \label{tab:instruments}
  \footnotesize\setlength{\tabcolsep}{3pt}
  \begin{tabular}{lp{2.6cm}p{3.8cm}}
    \toprule
    Code & Instrument & Source \\
    \midrule
    BFI  & Big Five Inventory & John, Donahue \& Kentle (1999); 44 items \\
    NFC  & Need for Cognition Scale & Cacioppo, Petty \& Kao (1984); 18-item short form \\
    CMS  & Consumer Minimalism Scale & Wilson \& Bellezza (2022) \\
    EMP  & Basic Empathy Scale & Carre, Stefaniak, D'Ambrosio, Bensalah \& Besche-Richard (2013) \\
    GRN  & Green Values Scale & Haws, Winterich \& Naylor (2014) \\
    CSC  & Conscientiousness Scale & Johnson, Vernon \& Feiler (2019), IPIP adjective form, 8 items \\
    SDS  & Social Desirability Scale & Reynolds (1982); 13-item short form \\
    RFS  & Regulatory Focus Scale & Fellner, Holler, Kirchler \& Schabmann (2007); 10 items \\
    MAX  & Maximization Scale & Nenkov, Morrin, Ward, Schwartz \& Hulland (2008); 6-item short form \\
    ACV  & Agentic vs.\ Communal Values Scale & Trapnell \& Paulhus (2012) \\
    ICS  & Individualism vs.\ Collectivism Scale & Triandis \& Gelfand (1998) \\
    NFU  & Need for Uniqueness Scale & Ruvio, Bagozzi \& Hult (2008); 12-item short form \\
    SMS  & Self-Monitoring Scale & Lennox \& Wolfe (1984); 13-item revised form \\
    NFCL & Need for Closure Scale & Roets \& Van Hiel (2011); 15-item short form \\
    SCC  & Self-Concept Clarity Scale & Campbell, Trapnell, Heine, Katz, Lavallee \& Lehman (1996); 12 items \\
    \bottomrule
  \end{tabular}
\end{table}

\section{Construct-Distance Tier Assignments}
\label{app:tiers}

Every unordered pair among the fifteen instruments above (105 pairs total) is assigned to one of the four tiers defined in Section~\ref{sec:distance}, by the family-based default and literature-backed override procedure given in full in Appendix~\ref{app:distance}; that procedure determines the tier for any given pair deterministically, so we summarize each tier by which instruments participate in it rather than enumerating all 105 individual pairs. The 139-pair grid in Section~\ref{sec:grid} uses a subset of these, expanded into ordered seed-to-measure runs (a pair can be run in one or both directions, and the Big Five Inventory is split into five separate facets when it is the measured instrument, which is why the counts here do not match the 6/52/75/6 ordered-pair tier counts reported there).

\textbf{Tier 1 (near-paraphrase), 2 pairs:} Big Five Inventory (BFI) with Conscientiousness Scale (CSC), and Agentic vs.\ Communal Values Scale (ACV) with Individualism vs.\ Collectivism Scale (ICS).

\textbf{Tier 2 (moderate distance), 30 pairs:} every instrument except Individualism vs.\ Collectivism Scale (ICS) participates in at least one tier-2 pair; the Big Five Inventory (BFI), Conscientiousness Scale (CSC), Social Desirability Scale (SDS), and Self-Monitoring Scale (SMS) are the most frequent partners.

\textbf{Tier 3 (substantial distance), 68 pairs:} all fifteen instruments participate; this is the largest and most diverse tier, and the default outcome for a pair with no closer family relationship and no literature override.

\textbf{Tier 4 (theoretically unrelated, negative control), 5 pairs:} all five pairs involve the Basic Empathy Scale (EMP), paired respectively with Consumer Minimalism Scale (CMS), Individualism vs.\ Collectivism Scale (ICS), Maximization Scale (MAX), Need for Uniqueness Scale (NFU), and Regulatory Focus Scale (RFS).

\section{Construct-Distance Methodology Detail}
\label{app:distance}

This appendix gives the full procedure summarized in Section~\ref{sec:distance}. Lexical overlap is computed at the item level: each item's text is lowercased, tokenized, stripped of a standard English stopword list, and lightly stemmed (common suffixes such as \emph{-ing}, \emph{-tion}, \emph{-ly}, and plural \emph{-s} are removed). For every item in instrument $a$ and every item in instrument $b$, we compute the Jaccard similarity of their token sets (the size of the intersection divided by the size of the union), and take the \emph{maximum} over all item pairs, not the mean, as that pair's lexical-overlap score; the maximum is deliberate, since a single near-duplicate item pair is enough to make a positive result attributable to surface repetition even if the remaining items are worded quite differently, and an average would understate that risk. Formally, for instruments $a$ and $b$ with item token sets $\{T^{a}_k\}$ and $\{T^{b}_l\}$,
\begin{equation}
  \mathrm{Lex}(a,b) = \max_{k,l} \frac{|T^{a}_k \cap T^{b}_l|}{|T^{a}_k \cup T^{b}_l|}.
\end{equation}
The construct-distance tier is assigned in two stages, so that most of the 105 pairs receive a defensible default while a documented minority receive a citable, specific judgment. First, each instrument is assigned to one of five thematic families (C.1): personality (the Big Five inventory), trait/cognitive-style scales, value and decision-style scales, a social-disposition family, and a response-bias family. A fixed table (C.2) gives the default tier for every family-by-family combination, reflecting the general expectation that two instruments from the same or a conceptually adjacent family are more likely to share real variance than two from unrelated families, absent a specific documented reason to think otherwise. Second, for pairs where a specific, citable empirical or theoretical relationship exists in the psychological literature, that relationship's tier assignment overrides the family default (C.3); an override always takes precedence, and family defaults apply only where no specific override has been identified.

\begin{table}
  \caption{Thematic family of each instrument.}
  \small\setlength{\tabcolsep}{3pt}
  \begin{tabular}{lll}
    \toprule
    Code & Instrument & Family \\
    \midrule
    BFI  & Big Five Inventory & Personality (Big Five) \\
    NFC  & Need for Cognition Scale & Trait / cognitive style \\
    CMS  & Consumer Minimalism Scale & Values / decision style \\
    EMP  & Basic Empathy Scale & Social disposition \\
    GRN  & Green Values Scale & Values / decision style \\
    CSC  & Conscientiousness Scale & Trait / cognitive style \\
    SDS  & Social Desirability Scale & Response bias \\
    RFS  & Regulatory Focus Scale & Values / decision style \\
    MAX  & Maximization Scale & Values / decision style \\
    ACV  & Agentic vs.\ Communal Values Scale & Values / decision style \\
    ICS  & Individualism vs.\ Collectivism Scale & Values / decision style \\
    NFU  & Need for Uniqueness Scale & Values / decision style \\
    SMS  & Self-Monitoring Scale & Trait / cognitive style \\
    NFCL & Need for Closure Scale & Trait / cognitive style \\
    SCC  & Self-Concept Clarity Scale & Trait / cognitive style \\
    \bottomrule
  \end{tabular}
\end{table}

\subsection{Family-pair default tiers}

\begin{table*}
  \caption{Default tier for each family-by-family combination.}
  \small\setlength{\tabcolsep}{3pt}
  \begin{tabular}{lc}
    \toprule
    Family pair & Default tier \\
    \midrule
    Personality (Big Five) $\leftrightarrow$ Trait / cognitive style  & 2 \\
    Personality (Big Five) $\leftrightarrow$ Values / decision style  & 3 \\
    Personality (Big Five) $\leftrightarrow$ Social disposition       & 2 \\
    Personality (Big Five) $\leftrightarrow$ Response bias            & 2 \\
    Trait / cog.\ style $\leftrightarrow$ Trait / cog.\ style         & 2 \\
    Trait / cog.\ style $\leftrightarrow$ Values / decision style     & 3 \\
    Trait / cog.\ style $\leftrightarrow$ Social disposition          & 3 \\
    Trait / cog.\ style $\leftrightarrow$ Response bias               & 2 \\
    Values / decision style $\leftrightarrow$ Values / decision style & 3 \\
    Values / decision style $\leftrightarrow$ Social disposition      & 4 \\
    Values / decision style $\leftrightarrow$ Response bias           & 3 \\
    Social disposition $\leftrightarrow$ Response bias                & 2 \\
    \bottomrule
  \end{tabular}
\end{table*}

\subsection{Literature-backed overrides}

An override applies to the specific pair named and supersedes the family-pair default above for that pair only. These are the overrides among the fifteen instruments used in this paper.

\begin{table*}[!htbp]
  \caption{Literature-backed tier overrides.}
  \footnotesize\setlength{\tabcolsep}{3pt}
  \begin{tabular}{p{3.3cm}p{3.6cm}cp{7.3cm}}
    \toprule
    Instrument A & Instrument B & Tier & Rationale \\
    \midrule
    Big Five Inventory (BFI) & Conscientiousness Scale (CSC) & 1 & Both directly operationalize the Big Five Conscientiousness facet; near-parallel measures of the same construct. \\
    Big Five Inventory (BFI) & Need for Cognition Scale (NFC) & 2 & Need for Cognition is the cognitive-engagement facet of Openness/Intellect (typical $r \sim .3$--$.4$ with BFI-O). \\
    Big Five Inventory (BFI) & Self-Monitoring Scale (SMS) & 2 & Self-monitoring's expressive-control items overlap with Extraversion/sociability. \\
    Big Five Inventory (BFI) & Self-Concept Clarity Scale (SCC) & 2 & Self-concept clarity loads negatively with Neuroticism / positively with emotional stability. \\
    Big Five Inventory (BFI) & Need for Closure Scale (NFCL) & 2 & Need for Closure correlates negatively with Openness ($r \sim -.3$ to $-.4$). \\
    Big Five Inventory (BFI) & Basic Empathy Scale (EMP) & 2 & Agreeableness is the Big Five facet most consistently linked to trait empathy ($r \sim .3$--$.4$). \\
    Big Five Inventory (BFI) & Social Desirability Scale (SDS) & 2 & Socially desirable responding correlates with Agreeableness/Conscientiousness (+) and Neuroticism ($-$) across the FFM literature. \\
    Need for Cognition Scale (NFC) & Need for Closure Scale (NFCL) & 2 & Need for Cognition and Need for Closure are reliably negatively correlated ($r \sim -.3$ to $-.4$) in the cognitive-style literature. \\
    Need for Closure Scale (NFCL) & Maximization Scale (MAX) & 2 & Maximizing and Need for Closure are linked constructs in the decision-style literature. \\
    Need for Closure Scale (NFCL) & Regulatory Focus Scale (RFS) & 2 & Need for Closure aligns with prevention-focused regulatory orientation (both emphasize certainty over exploration). \\
    Agentic vs.\ Communal Values Scale (ACV) & Individualism vs.\ Collectivism Scale (ICS) & 1 & Agency/Communion is presented in its own literature as the individual-difference analogue of Individualism/Collectivism's self-construal dimension. \\
    Consumer Minimalism Scale (CMS) & Green Values Scale (GRN) & 2 & Consumer minimalism and pro-environmental values are consistently positively correlated in sustainable-consumption research. \\
    Consumer Minimalism Scale (CMS) & Need for Uniqueness Scale (NFU) & 2 & Both concern distinctiveness-motivated consumption restraint/selection, though valence differs (avoiding excess vs.\ conformity). \\
    Regulatory Focus Scale (RFS) & Maximization Scale (MAX) & 2 & Regulatory focus theory is explicitly invoked in the maximizing/satisficing decision literature. \\
    Green Values Scale (GRN) & Basic Empathy Scale (EMP) & 2 & Dispositional empathy predicts pro-environmental concern in environmental-psychology research (`empathy for nature'). \\
    Agentic vs.\ Communal Values Scale (ACV) & Basic Empathy Scale (EMP) & 2 & The Communion pole is defined by caring/altruism/compassion items overlapping with trait empathy's affective-concern facet. \\
    Green Values Scale (GRN) & Social Desirability Scale (SDS) & 2 & Pro-environmental self-report is a frequently cited domain for social-desirability contamination. \\
    Basic Empathy Scale (EMP) & Social Desirability Scale (SDS) & 2 & Self-reported empathy is itself socially desirable to claim; documented positive correlation with social-desirability responding. \\
    \bottomrule
  \end{tabular}
\end{table*}

\end{document}